%% file: ITSC2021.tex
\documentclass[twocolumn]{article}
\usepackage{arxiv}

\usepackage[utf8]{inputenc}
\usepackage[english]{babel}
\usepackage{multicol}
\usepackage{float}
\usepackage{bbm}
\usepackage{amsmath,amsfonts}
\usepackage{multirow}
\usepackage{array}
\usepackage{graphicx}
\usepackage{booktabs}
\usepackage[numbers,sort&compress,square]{natbib}
\usepackage{xcolor}
\usepackage{tabularx}
\graphicspath{{./images/}}

\usepackage{blindtext}

\usepackage{amsmath}
\usepackage{autobreak}

\usepackage{enumitem}
\setlist[itemize]{noitemsep,topsep=3pt,leftmargin=*}

\usepackage[original]{abstract} %

\usepackage{balance}

\usepackage[utf8]{inputenc}
\usepackage[T1]{fontenc}
\usepackage{amsmath}
\usepackage{bbm}
\usepackage{lipsum}
\usepackage{graphicx}
\usepackage{mathtools, nccmath}
\usepackage[ruled]{algorithm2e}
\usepackage{algpseudocode}
\SetKwInput{KwInput}{Input}                %
\SetKwInput{KwOutput}{Output}
\usepackage{siunitx}

\usepackage{booktabs}
\usepackage[numbers,sort&compress,square]{natbib}

\usepackage[original]{abstract} %

\usepackage{graphics} %
\usepackage{epsfig} %
\usepackage{subfig}
\usepackage{algorithmicx} 
\title{Navigation in Urban Environments amongst pedestrians using Multi-Objective Deep Reinforcement Learning}

\usepackage{blindtext}

\usepackage{amsmath}
\usepackage{autobreak}

\usepackage{enumitem}
\setlist[itemize]{noitemsep,topsep=3pt,leftmargin=*}

\author{Niranjan Deshpande$^{1}$, Dominique Vaufreydaz$^{2}$ and Anne Spalanzani$^{1}$%
\thanks{$^{1}$ Univ. Grenoble Alpes, Inria, 38000 Grenoble, France, {\tt\small email: FirstName.LastName@inria.fr}}%
\thanks{$^{2}$Univ. Grenoble Alpes, CNRS, Grenoble INP, LIG, 38000 Grenoble, France}%
}

\begin{document}

\twocolumn[{%
  \begin{@twocolumnfalse}
    \maketitle
  \end{@twocolumnfalse}
  
}]

\saythanks 
\setcounter{footnote}{0}

\setlength{\abovedisplayskip}{3pt}
\setlength{\belowdisplayskip}{3pt}


\begin{abstract}
Urban autonomous driving in the presence of pedestrians as vulnerable road users is still a challenging and less examined research problem.   
This work formulates navigation in urban environments as a multi objective reinforcement learning problem.
A deep learning variant of thresholded lexicographic Q-learning is presented for autonomous navigation amongst pedestrians.
The multi objective DQN agent is trained on a custom urban environment developed in CARLA simulator.
The proposed method is evaluated by comparing it with a single objective DQN variant on known and unknown environments. 
Evaluation results show that the proposed method outperforms the single objective DQN variant with respect to all aspects.
\end{abstract}

\input{introduction}

\input{background}

\input{related_work}

\input{problem_statement}

\input{method}

\input{results}

\input{conclusion}

\section{ACKNOWLEDGMENT}
This work was funded under project CAMPUS (Connected Automated Mobilty Platform for Urban Sustainability) sponsored by Programme d'Investissements d'Avenir (PIA) of french Agence de l'Environnement et de la Ma\^itrise de l'\'Energie (ADEME).

\input{bibliography}
\end{document}

%% file: introduction.tex
\section{INTRODUCTION}

One of the most important aspects of autonomous driving is navigation. A typical navigation system of an autonomous vehicle employs a sequential pipeline which is widely used by the industry and is considered the conventional approach \cite{c1}, \cite{c2}. It consists of a \textit{Route Planner} at the highest level, generating a list of sparse way-points from start to destination. 
This is followed by a \textit{Behavioral Decision  Making} (BDM) layer,
responsible for deciding on a driving maneuver based on the information of other traffic participants' behavior, road conditions and traffic signals.
The minimal goal of BDM is to ensure safety, while the other goals are speed, comfort, fuel efficiency. Based on the selected behaviour, \textit{Motion Planning} and \textit{Control} modules are used for continuous path generation and execution respectively.

A natural approach to automating the BDM is to model each of its behavior as states in a finite state machine (FSM). In fact, FSMs coupled with different heuristics were adopted as a mechanism for behavior planning by most teams in the DARPA challenges, given their simplicity and transparency. However, their major drawback is the difficulty to model the uncertainties and complex urban traffic scenarios. 

In recent years, Deep Reinforcement Learning (deep-RL), has been successfully applied to several sequential decision-making problems. 
This has let to a surge in the use of deep-RL based decision making approaches for autonomous driving as well.
However, these approaches are applied for highway driving scenarios in the presence of vehicles as other road users. 

Furthermore, existing deep-RL approaches for autonomous driving consider navigation as a simplified single objective RL problem and employ a scalar reward function for different objectives like safety and speed.
This simplification works for highway driving, however, in the context of urban environments, an autonomous vehicle often has to satisfy several conflicting objectives: 
avoiding collision with vulnerable road users such as pedestrians, maintaining speed limits, following traffic rules and ensuring passenger comfort.

This work formulates autonomous navigation in urban environments as a multi-objective RL problem where each objective is learned by a separate agent. 
A common policy is then formulated by these agents by taking into consideration all these objectives.
The motivation behind this rationale is the difficulty in designing a scalar reward that could properly weigh the importance of each of the above mentioned objectives.

This work presents the following contributions:
\begin{itemize}
    \item A multi-objective RL method called thresholded lexicographic Deep Q-learning is presented for navigation in urban environments in the presence of pedestrians.
    \item A single-objective RL method presented in our previous work \cite{c6, c7}, is improved with a new reward function and used for comparison purposes.
    \item For training, a custom urban environment is developed and the proposed methods are compared and evaluated to quantify their performance.
\end{itemize}

%% file: background.tex
\section{BACKGROUND}
\subsection{Single Objective Reinforcement Learning}
Reinforcement learning (RL) is often formulated as a Markov Decision Process (MDP), a tuple \textit{(S, A, T, R)}, characterized by a set of states \textit{S}, a set of actions \textit{A} and a scalar reward function \textit{R} and a transition function \textit{T}. 
In the single-objective reinforcement learning (SORL), 
at each time \textit{t}, an agent interacts with the environment by observing the state \textit{s\textsubscript{t}} and performing an action \textit{a\textsubscript{t}}, according to some policy $\pi$. Performing an action generates a reward \textit{r\textsubscript{t}} and takes the agent to next state \textit{s\textsubscript{t+1}} and the process repeats until a termination condition.
The agents goal is to maximize the expected discounted
reward $R_{t} = \sum_{t=0}^{T} \gamma \cdot r_{t}$, where $\gamma \in [0, 1]$ is a discount factor deciding the importance of future rewards.

In Q-learning algorithm, actions are selected using a Q-function \textit{Q(s,a)}, 
which constitutes the expected discounted reward for executing an
action \textit{a} in state \textit{s}.
The optimal action for a state \textit{s} is defined as:
\begin{equation}
    a_{}{'} = \underset{a \in A}{\arg\max}\:Q(s, a) 
\end{equation}
An agent is said to be following an optimal policy $\pi^{*}$, given it always selects the optimal action. The function $Q(s,a)$ helps in forming an optimal policy by simply selecting actions with highest Q-values.

In Deep Q-Networks, the $Q(s,a)$ values are approximated using deep neural networks  making Q-learning feasible for many real-world complex applications.
A variant of DQN, called Double Deep Q-Network (DDQN) uses a separate target network to make the learning process more stable. Its parameters are periodically updated using actual network.

\subsection{Multi-Objective Reinforcement Learning}
In several complex applications, there exists more than one objective to be considered.
If these objectives are independent or directly related, they could be merged to form a single objective. However, often the objectives are in conflict, that is, maximizing one objective causes minimization of another.
For such applications, multi-objective reinforcement learning (MORL) is used to characterize two or more objectives simultaneously.
MORL is often modelled as Multi-objective Markov Decision Process (MOMDP) instead of single objective MDP.
An MORL agent, instead of a scalar reward, receives a vector of rewards at each time step, corresponding to each objective. 
Given \textit{O} as a set of \textit{n} objectives, the reward vector is defined as: 
\textit{R} $=$ [\textit{r\textsubscript{1}}, \textit{r\textsubscript{2}},...,\textit{r\textsubscript{n}}], where \textit{r\textsubscript{i}} represents scalar reward for objective \textit{o\textsubscript{i}}$\in$\textit{O}.
The expected discounted reward for an objective \textit{o\textsubscript{i}}, at time \textit{t}, is defined as:
\useshortskip
\begin{equation}
    R_{i,t} = \sum_{t=0}^{T} \gamma \cdot r_{i,t}
\end{equation}
\useshortskip
Usually, a separate Q-function $Q_{i}(s,a)$ is considered for each objective such that:
\begin{equation}
    Q_{i}(s,a) = E [R_{i,t}|s_{t} = s, a_{t} = a]
\end{equation}
resulting into a vector of Q-functions.
At each time step \textit{t}, the agent uses the function $Q_{i}(s,a)$ to find the optimal action satisfying objective \textit{o\textsubscript{i}}.
Since the agent can perform a single action at a time, a truncation method is required.

\subsection{Thresholded Lexicographic Q-learning}
Threshold lexicographic Q-Learning (TLQ-Learning), introduced by \textit{Gabor et al.} \cite{c8} is an algorithm which uses lexicographic ordering on objectives based on their priorities.
A threshold is used for each objective to define the minimum or
maximum acceptable Q-value (depending on whether the objective is
to be minimised or maximised).
Thresholds are set on the first $i-1$ objectives and the last objective
is left unconstrained.

Selection of action is done by using a combination of thresholding and lexicographic ordering to the objective's Q-values. 
If $Q_{i}(s,a)$ is the value for \textit{i\textsuperscript{th}} objective corresponding to action \textit{a} in state \textit{s} and \textit{T\textsubscript{i}} is threshold, then:
\begin{equation}
    TQ_{i}(s,a) = min(Q_{i}(s,a), T_{i})
\end{equation}
Given a state \textit{s}, a greedy action $a_{}{'}$ is selected, satisfying the recursive function as defined in Algorithm \ref{algorithm:tlo} \cite{c9}.
\input{TLO_algorithm}

%% file: TLO_algorithm.tex
\begin{algorithm}[t]
    \SetKwInOut{Input}{Input}
    \SetKwInOut{Output}{Output}

    \Input{$TQ(s,a)$, $TQ(s,a^{'})$, \textit{i}}
    \Output{if $TQ(s,a)$ superior to $TQ(s,a^{'})$}
    \uIf{$TQ_{i}(s,a) > TQ_{i}(s,a^{'})$}
    {
        \textbf{return} true\;
    }
    \uElseIf{$TQ_{i}(s,a) = TQ_{i}(s,a^{'})$}
    {
        \uIf{i = n}
        {
            \textbf{return} true
        }
        \Else
        {
            \textbf{return} Superior($TQ(s,a)$, $TQ(s,a^{'})$, $i+1$)    
        }
    }
    \Else
    {
        \textbf{return} false
    }
    \caption{Superior}
    \label{algorithm:tlo}
\end{algorithm}

%% file: related_work.tex
\section{RELATED WORK}
\subsection{Navigation in the presence of vehicles}
In recent years, deep-RL for navigation on highway environments has been explored extensively. Several approaches use Deep Q-Network (DQN) and its variants for lane change decision making. 
A detailed discussion of relevant approaches could be found in our previous work \cite{c6, c7} and in survey papers \cite{c1, c2}.

Overtaking maneuver is formulated as a multi-objective reinforcement learning problem by \textit{Ngai and Yung} in \cite{c10}. Values from a tabular Q-function of each objective are scalarized by weighted sum into a single policy.  
Another approach is proposed by \textit{C. Li and K. Czarnecki} in \cite{c11} for urban intersection handling. 
The trained multi-objective DQN agent learns to drive on multi-lane roads and intersections, yielding and changing lanes according to traffic rules. It presents a deep learning variant of thresholded lexicographic Q-learning for the task of autonomous driving.

\subsection{Navigation in the presence of pedestrians}
There exists some prior work on navigation of small mobile robots around pedestrians in outdoor environments.
\textit{Bai et al.} in \cite{c12} propose an intention aware method for navigation in environments with pedestrians. A hybrid $A^{*}$ planner is used for global path planning, while an online POMDP (Partially observable Markov decision process) is used for velocity control.
An intention awake decision maker is proposed in \cite{c13} where pedestrian behavior is modelled in terms of potential fields.
An POMDP approach in \cite{c14} proposes a pedestrian motion predictor called PORCA to model their intentions and interactions.
\textit{Barbier  et  al.} in \cite{c15} model autonomous vehicle and pedestrian interaction at intersections using POMDP.  
However, solving a POMDP exactly is an intractable problem. 
Safety of POMDP solutions in the context of autonomous driving is still a topic of research \cite{c18}. Furthermore, online POMDP planners are often computationally complex and require much time, limiting their use for autonomous vehicle platforms.

Some approaches also explore the use of game theory. 
In \cite{c16}, \textit{Fox et al.} propose a game-theoretical mathematical model called \textit{sequential chicken}, for two agents meeting at an unsigned interaction and negotiating for priority. 
The \textit{sequential chicken} model is further extended in \cite{c17} to negotiate collision with a pedestrian at an unsignalized intersection.
However, game-theoretic approaches are too simplistic to model complex autonomous vehicle's interactions with pedestrians \cite{c18}.

In our previous work \cite{c6, c7}, a unique 3-D grid representation, suitable for urban environments, is used by a DQN agent to learn navigation in environment densily populated with pedestrians.
This 3-D representation is also evaluated by \textit{K. Mokhtari} and \textit{A. Wagner} with different variants of DQN \cite{c18}.
They further propose a safe deep-RL approach in \cite{c19} where an LSTM based future collision algorithm is used to mask unsafe actions.

%% file: problem_statement.tex
\section{PROBLEM STATEMENT}
Autonomous navigation is usually formulated as a single-objective RL problem, especially for highways.
However, in the context of urban environments, navigation is more challenging, since along with other vehicles, there are vulnerable road users such as pedestrians and cyclists.
Pedestrian motions are less constrained and even slightest collision is likely to be fatal. 
Furthermore, an autonomous vehicle navigating in urban setups often needs to satisfy multiple conflicting objectives simultaneously, such as: collision avoidance, speed control, traffic rules and passenger comfort.

This work formulates autonomous navigation in urban environments as a multi-objective RL problem, where each objective is learned separately and a combined policy is formed that takes into account all the objectives. These objectives have different priorities associated with them.
For instance, an autonomous vehicle needs to avoid any collision situations before considering speed control and traffic rules.
Since thresholded lexicographic learning follows a similar procedure, this work extends the deep learning variant of thresholded lexicographic Q-learning proposed in \cite{c11} for the task of urban autonomous driving amongst pedestrians.

%% file: method.tex
\section{METHODOLOGY}
This work considers an autonomous vehicle as an RL agent with multiple objectives, receiving rewards with respect to these objectives and having a separate Q-function for each of its objectives. 

\subsection{Objectives}
The agent should avoid collisions with surrounding pedestrians and simultaneously drive at the desired speed.
For this, two objectives in lexicographic order are considered:
\begin{itemize}
    \item safety (\textit{o\textsubscript{safety}}): To avoid any collision situations. 
    \item speed (\textit{o\textsubscript{speed}}): To ensure the agent drives as far as possible, within the speed limits.
\end{itemize}
\subsection{MOMDP formulation}
\subsubsection{State Space}
The state space contains information about the surrounding environment and the ego vehicle itself.

To store environment information (\textit{environment state}), a multi-layered 3-D grid representation \cite{c6} is employed by defining a region of interest (ROI) around the ego vehicle with length L and width W.
This ROI is further discretized into multiple 2-D grids, with each grid storing a particular type of information: occupancy, relative  speed (in \SI{}{\meter/\second}), relative heading (in degrees) and semantic information (road type).

For ego vehicle state information (\textit{ego vehicle state}), its current velocity (in \SI{}{\meter/\second}) is available.

\subsubsection{Reward Vector} \label{reward_vector}
In MORL paradigm, the reward for the agent is in terms of a vector corresponding to each objective. In this work the following reward vector is considered:
\begin{equation}
    R = [\textit{r\textsubscript{safety}}, \textit{r\textsubscript{speed}}]
\end{equation}
where \textit{r\textsubscript{safety}} is the reward associated with the \textit{o\textsubscript{safety}} objective and \textit{r\textsubscript{speed}} is for \textit{o\textsubscript{speed}}.
\paragraph{Safety reward}
With respect to safety, an autonomous vehicle should be able to avoid any collisions while driving. After testing several reward combinations, the following function is used to formulate the reward:
  \begin{equation}
    \textit{r\textsubscript{safety}} =
    \begin{cases}
      r_{c} & \textit{for collision}\\
      r_{nc} & \textit{for near collision}\\
    \end{cases}
  \end{equation}
where $r_{c}= -4$, is a large collision penalty and $r_{nc}$ is a penalty for near collision situations. To compute $r_{nc}$, a dynamic distance $(d_{r})$ range is defined as a function of ego-vehicle's current speed $(v_{ev})$. If there exists a pedestrian crossing the road within $(d_{r})$, a penalty for the nearest front pedestrian is calculated using the following equations \cite{c20}:
\begin{align}
\label{equation:near_collision}
\begin{split}
 & d_{r} = max\left( \frac{v_{ev}^2}{2a_{max}}, d_{0} \right) \\
 & r_{nc} = exp\left(  \frac{d_{p} - d_{r}}{d_{r}} \right)
\end{split}
\end{align}
where, $a_{max}$ denotes the ego-vehicle's maximum deceleration, $d_{0}$ is a minimum safety distance to the front pedestrian and $d_{p}$ is the distance to the  pedestrian.
\paragraph{Speed reward}
The reward for speed is computed using the following function:
  \begin{equation}
    \textit{r\textsubscript{speed}} =
    \begin{cases}
      \lambda\cdot(v_{ref} - v_{ev}), & \textit{if} \quad \text{0.0} < v_{ev} \leq v_{ref}\\
      -1.0, & \textit{if}\quad{ v_{ev}} \leq \text{0.0}\\
      -0.5, & \textit{if}\quad v_{ev} > v_{ref}\\
    \end{cases}
  \end{equation}
where $\lambda = 1/v_{ref}$, $v_{ev}$ is ego vehicle current speed and $v_{ref}$ is the target desired speed.

\subsubsection{Action Space} \label{action_space}
\sisetup{per-mode=symbol}
The action space consists of four high level actions for each DQN agent.
\begin{center}
\begin{tabular}{ |l|l|c| } 
\hline
\textbf{Actions} & \textbf{Description} & \textbf{Values} \\ 
\hline
\textit{accelerate} & Gradual speed-up & +\SI{1}{\metre\per\second\squared} \\ 
\hline
\textit{decelerate} & Gradual slow-down & -\SI{1}{\metre\per\second\squared} \\ 
\hline
\textit{brake} & Braking & -\SI{5}{\metre\per\second\squared} \\
\hline
\textit{steer} & Maintain current speed & \SI{0}{\metre\per\second\squared} \\
\hline
\end{tabular}
\end{center}

 \subsection{Thresholded Lexicographic DQN}
 A separate DQN is used as an approximator for the Q-function of each objective: 
 \textit{DQN\textsubscript{safety}} and \textit{DQN\textsubscript{speed}}.
 Each DQN receives a different state and reward with respect to their objectives, however, both have same action space (described in \ref{action_space}).
 The \textit{DQN\textsubscript{safety}} uses \textit{environment state} as its state space input and \textit{r\textsubscript{safety}} is the reward generated. For \textit{DQN\textsubscript{speed}}, the state is \textit{ego vehicle speed} and reward generated is \textit{r\textsubscript{speed}}.
 
In single objective RL, given a state \textit{s}, the algorithm selects the greedy action $a_{}{'}$ as optimal action, such that:
\begin{equation}
    a_{}{'} = \underset{a \in A}{\arg\max}\:Q(s, a)
\end{equation}
Whereas, Thresholded Lexicographic Q-learning (TLQ) selects an action for each objective based on a threshold that specifies the minimum admissible value for each objective. 
However, this work uses a slightly different meaning of the threshold, specifying how much worse than the optimal action is considered acceptable \cite{c11}.
At every time step, a list of acceptable actions $(A_{i}{\in}{A})$ is generated for each objective, however, the allowable actions for \textit{i\textsuperscript{th}} objective are restricted to those allowed by objective \textit{i-1}.

The threshold is defined in terms of percentage:
\begin{equation}
    T = [\tau_{safety}, \tau_{speed}]
\end{equation}
For example, at time t, the optimal action by \textit{DQN\textsubscript{safety}} is $a_{}{'}$, then all other actions whose Q-value is greater than or equal to $\tau_{safety}*Q(s, a_{}{'})$ are considered acceptable for objective \textit{o\textsubscript{safety}}.
Then out of these acceptable actions, \textit{DQN\textsubscript{speed}} selects the optimal action for objective \textit{o\textsubscript{speed}} based on its state input and learned policy.
The action selection procedure is described in Algorithm \ref{algorithm:tlo_action_selection} \cite{c11}. The input to the algorithm is a list of learned Q-functions for each objective, a list of states for each objective and a boolean list indicating which objective is to be explored.  
\input{TLO_action_selection}
\vspace{-0.5cm}

%% file: TLO_action_selection.tex
\newcommand\mycommfont[1]{\footnotesize\ttfamily{#1}}
\SetCommentSty{mycommfont}
\begin{algorithm}
\SetNoFillComment
\caption{TLO Action Selection}
    \SetKwInOut{Input}{Input}
    \SetKwInOut{Output}{Output}

    \Input{vector Q, vector S, vector E}
    \Output{final action}
    $A_{0} := A$ \\
    \ForEach{i $\in$ n}
    {
        \begin{multline*}
        A_{i}(s_{i}) := \\
        \Biggl\{ a\in A_{i-1}(s_{i-1})\Big|Q_{i}(s_{i},a)\geq \underset{a_{}{'}\in A_{i-1}}{\max}Q_{i}(s_{i}, a_{}{'})*\tau_{i}\\
        \text{\textbf{or}}\quad a = \underset{a_{}{'}\in A_{i-1}}{\arg\max}\:Q_{i}(s_{i},a_{}{'}) \Biggr\} 
        \end{multline*}
    
        \tcc{if objective i chosen to be explored}
        \If{E[i] == True}{
            \Return random action from $A_{i-1}(s_{i-1})$
        }
    }
    \Return greedy action from $A_{n}(s_{n})$
    \label{algorithm:tlo_action_selection}
\end{algorithm}

%% file: results.tex
\section{EXPERIMENTS}
\subsection{Neural Network Models}\label{neural_network_models}
\begin{figure}[htp]
    \centering
    \includegraphics[width=\linewidth, height=6.2cm]{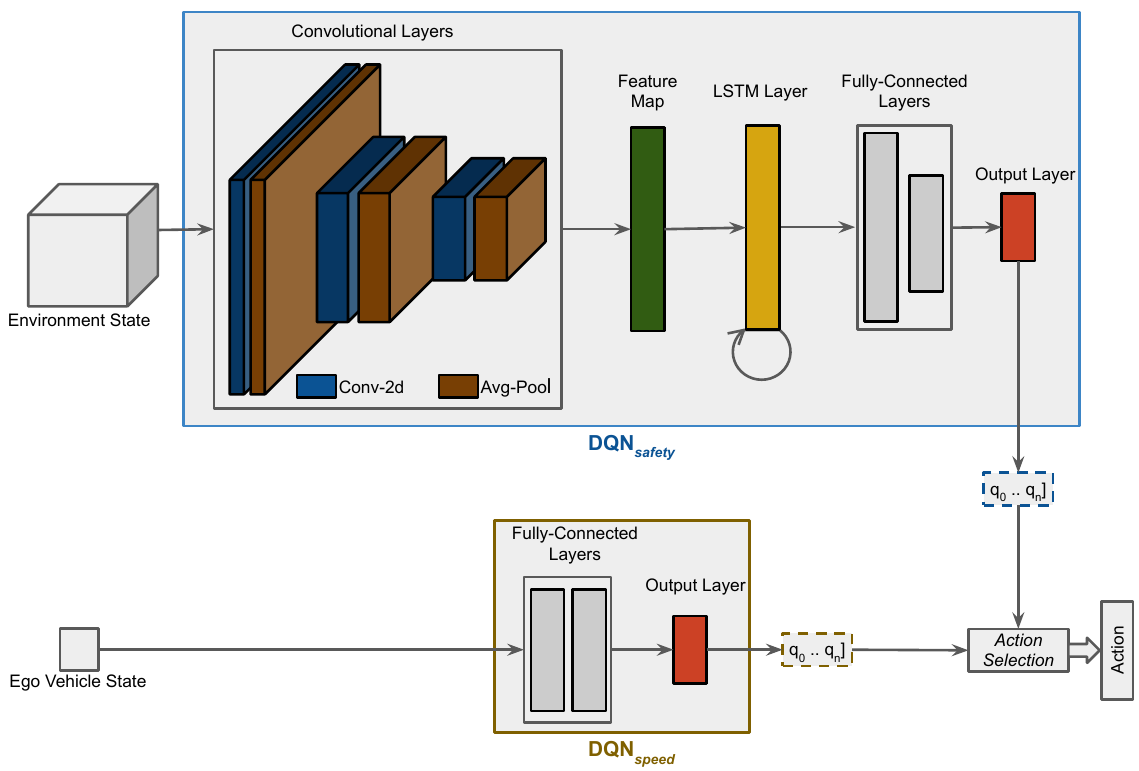}
    \caption{Separate DQNs used for \textit{o\textsubscript{safety}} and \textit{o\textsubscript{speed}} objectives. Thresholding and lexicographic ordering is performed on the Q-value outputs from these DQNs to generate final action}
    \label{figure:lstm-nn}
\end{figure}
Two different MORL agents are investigated in this work, labelled as \textit{Agent1} and \textit{Agent2} respectively.
The difference lies in the neural network (NN) architecture of their \textit{DQN\textsubscript{safety}}, while both agents have the same architecture for their \textit{DQN\textsubscript{speed}}.

\paragraph{Architecture for \text{DQN\textsubscript{safety}}}
The input to \textit{DQN\textsubscript{safety}} is the multi-layered 3-D grid
of size $80\times60\times4$. 
The input is passed through three convolutional layers.
The first layer has \textit{n\textsubscript{conv1}} filters set to 32, 
second layer has \textit{n\textsubscript{conv2}} filters set to 64,
third layer has \textit{n\textsubscript{conv3}} filters set to 64.
Each layer uses a filter size of 5, stride of 3, has a ReLU activation function and is also followed by an AveragePooling layer for dimentionality reduction.

For \textit{Agent1}, the output from these convolutional layers is fed into two fully connected layers with \textit{n\textsubscript{full1}} and \textit{n\textsubscript{full2}} units, set to 128 and 64 respectively, followed by ReLu activation functions.  Finally, the output layer has 4 neurons corresponding to the action set.

For \textit{Agent2}, the output from the convolutional layers is flattened and fed into an LSTM layer as shown in Figure \ref{figure:lstm-nn} with \textit{n\textsubscript{lstm}} hidden units, set to 128. The output of LSTM is then fed into fully connected layers followed by the output layer.

\paragraph{Architecture for \text{DQN\textsubscript{speed}}}
Both the agents use same NN design for their \textit{DQN\textsubscript{speed}}.
The input is the \textit{ego vehicle state} (its current speed) which is passed through 
two fully connected layers with \textit{n\textsubscript{full1}} and \textit{n\textsubscript{full2}} units, set to 32 and 32, followed by ReLu activation. The output layer is of size 4.

\subsection{Training Details}
For the purpose of training, CARLA simulator \cite{c21} is used.
Both the agents are trained for a total of 500000 steps. 
However, an episode ends if the agent reaches the goal or encounters a collision situation.

Both the DQNs maintain a separate experience replay memory to store up to 10,000 transitions.
A mini-batch of size 32 is selected uniformly from the replay memory buffer by each DQN and the network weights are then updated accordingly using RMSProp algorithm \cite{c22}.
Learning rate for \textit{DQN\textsubscript{safety}} is set to 0.00025, while for \textit{DQN\textsubscript{speed}} its 0.0025 respectively.
Both the DQNs are trained using the Double-DQN algorithm with target networks which are updated every 1000 steps.
However, since \textit{DQN\textsubscript{safety}} of \textit{Agent2} has an LSTM layer, it uses a mini-batch of 32 experiences, each experience consisting of 4 timesteps. 

At each time step, one of the objectives is randomly selected for exploration. 
If objective \textit{o\textsubscript{safety}} is chosen for exploration. then objective \textit{o\textsubscript{speed}} is no longer considered for action selection (described in Algorithm \ref{algorithm:tlo_action_selection}). 
Each DQN uses an $\epsilon$ greedy approach for action selection, where the action is selected randomly with a probability of $\epsilon$, else an action with the highest Q-value is selected.
For \textit{DQN\textsubscript{safety}}, the $\epsilon$ is initialized to 0.9 and it linearly decreases to a minimum value of 0.3 over 4,00,000 steps.
For \text{DQN\textsubscript{speed}}, $\epsilon$ decreases from 0.8 to 0.1  
linearly over 4,00,000 steps.

\subsection{Simulation Details}
The proposed method is trained and evaluated in CARLA simulator \cite{c21}, an open source simulator for autonomous driving.  
A custom urban environment consisting of two-way roads, four-way unsignalized intersections with crosswalks is developed in the simulator.  
The simulator provides the environment and ego vehicle information and receives desired speed values for the ego vehicle as input.  
The simulator time step is set at $t=0.1 sec$. 

At the beginning of each episode, the ego vehicle is spawned at the start position and follows a list of way-points generated by the simulator to reach its destination. 
The desired speed limit for the ego vehicle is set at \SI{8}{\metre\per\second}.

At each time step, a total of up to 30 pedestrians exists within a vicinity of 35 meters around the ego vehicle (in front) with random goals and velocities between \SI{0.4}{}-\SI{1.2}{\metre\per\second}.
A pedestrian crossing factor (set to 0.8) is used, indicating how many pedestrians can cross the road (80\%).
Pedestrians moving farther away (40 meters) from the ego vehicle are removed and an equal number of new pedestrians are spawned to make sure busy pedestrian traffic exists around the ego vehicle.

\subsection{Results}
As described in Section \ref{neural_network_models}, two MORL agents with different neural network architectures are investigated in this work. \textit{Agent1} uses convolutional layers followed by fully connected layers for \textit{DQN\textsubscript{safety}}.
However, \textit{Agent2} adds a LSTM layer after the convolutional layers adding memory to the \textit{DQN\textsubscript{safety}} network. Both agents use same network structure for their \textit{DQN\textsubscript{speed}}.

Furthermore, the proposed MORL agents are compared with two SORL agents as well, which are based on our previous work in \cite{c7} but with an improved reward function.
Both these SORL agents use a single DQN to satisfy both the objectives and learn the desired policy.
\textit{Agent1} of SORL has a similar neural network architecture as that of MORL \textit{Agent1}, where the \textit{environment state} (3-D multilayered grid) information is passed through convolutional layers followed by fully connected layers. However, the \textit{ego vehicle state} (its current speed) information is concatenated with the first fully connected layer and passed to the output layer.
\textit{Agent2} of SORL, uses two separate LSTMS: one for the \textit{environment state} and other for \textit{ego vehicle state}. The \textit{environment state} is passed through convolutional layers to the first LSTM and the \textit{ego vehicle state} is directly passed to the second LSTM. Outputs of both LSTMs are then concatenated and passed to fully connected layers followed by the output layer.
Since a SORL agent expects reward as a scalar number, the safety (\textit{r\textsubscript{safety}}) and speed (\textit{r\textsubscript{speed}}) rewards described in Section \ref{reward_vector} are added up and passed as a single reward value to the SORL agents.
In summery, four variants are compared, two SORL agents (\textit{Agent1\textsubscript{CNN}}, \textit{Agent2\textsubscript{LSTM}}) and two proposed MORL agents (\textit{Agent1\textsubscript{CNN}}, \textit{Agent2\textsubscript{LSTM}}).

All the agents are tested for 100 episodes in the known environment (training environment) and on unseen new environments. 
\textit{Town01} and \textit{Town04} of CARLA simulator are used as unseen environments.
\textit{Town01} is similar to the custom training environment but consists of three-way intersections.
While \textit{Town04} is an urban environment with four-way roads. 
The agents are compared and evaluated using the following metrics:
\begin{itemize}
    \item \textbf{Collision free episodes: }Percentage of episodes executed without collision.
    \item \textbf{Success rate: }Percentage of episodes where agent reached the goal in time.
    \item \textbf{Distance travelled: }Average distance travelled during each episode.
    \item \textbf{Average Steps: }Number of steps executed in each episode.
    \item \textbf{Average speed: }Average speed of the agent during each episode.
    \item \textbf{Speed violation: }Percentage of episodes where the target speed limit was violated by the agent.
    \item \textbf{Average crossing duration: } This indicates how long it takes for the agent to handle an intersection, calculated in terms of percentage of steps at the intersection with respect to total steps.
    \item \textbf{Average Stops: }Number of times the agent brakes to stop in every episode.
    \item \textbf{Distance to closet pedestrian: }Average distance to closest ($<2m$) front crossing pedestrian in the same lane.
\end{itemize}
\input{results_table}

\subsection{Discussion}
The test results for all the agents on seen and unseen environments are presented in Table \ref{results}. The proposed MORL agents clearly outperforms the SORL agents with respect to each measure. As mentioned in the above sections, collision avoidance is the foremost objective of an autonomous vehicle driving in urban environments amongst pedestrians.
The SORL agents already perform better with 97\% and 98\% collision free episodes, however, 
the MORL method shows further improvements with 100\% collision free episodes for both the agents in known as well as unknown environments. 
Another interesting aspect is the number of times braking is performed. The SORL performs more braking to stop compared to MORL. This is because SORL agents use a scalar reward and are always trying to balance the penalties/rewards received for near collisions and ego vehicle speed. However, the MORL agents perform less stopping actions resulting in more natural driving.  
The MORL agents show 100\% success rates in known environments, but in unknown environments their success rate is 98\%. All the agents avoid any speed violations indicating that the ego vehicle speed is always within the speed limits. However, agents with a LSTM layer maintain higher speeds since they have memory and can anticipate the intentions of surrounding pedestrians better. The advantage of having a memory is also reflected in terms of lesser waiting times at the intersections and higher pedestrian distance maintained by LSTM based agents.

%% file: results_table.tex
\begin{table*}[ht]
 \centering
\resizebox{\textwidth}{!}{\begin{tabular}{l cccc cccc}
\toprule
Environment & \multicolumn{4}{c}{Known Environment} & \multicolumn{4}{c}{Unknown Environments} \\
\cmidrule(lr){2-5} \cmidrule(lr){6-9}

Agent & \multicolumn{2}{c}{Single Objective} & \multicolumn{2}{c}{Multi Objective} & \multicolumn{2}{c}{Single Objective} & \multicolumn{2}{c}{Multi Objective} \\

Architecture & \multicolumn{1}{c}{\textit{Agent1\textsubscript{CNN}}} & \multicolumn{1}{c}{\textit{Agent2\textsubscript{LSTM}}} & \multicolumn{1}{c}{\textit{Agent1\textsubscript{CNN}}} & \multicolumn{1}{c}{\textit{Agent2\textsubscript{LSTM}}} & \multicolumn{1}{c}{\textit{Agent1\textsubscript{CNN}}} & \multicolumn{1}{c}{\textit{Agent2\textsubscript{LSTM}}} & \multicolumn{1}{c}{\textit{Agent1\textsubscript{CNN}}} & \multicolumn{1}{c}{\textit{Agent2\textsubscript{LSTM}}}\\
\midrule
Collision Free Episodes (\%) & 97\% & 98\% & \textbf{100}\% & \textbf{100}\% & 96\% & 98\% & \textbf{100}\%\ & \textbf{100}\% \\
Success Rate (\%) & 95\% & 98\% & \textbf{100}\% & \textbf{100}\% & 95\% & 97\% &\textbf{98}\% & \textbf{98}\% \\
Distance Travelled (m) & 142.48 & 145.1 & \textbf{152.1} & \textbf{152.1} & 118.1 & 118.96 & \textbf{122.01} & \textbf{123.0} \\
Average Steps & 779.62 & 603.4 & \textbf{721.1} &\textbf{598} & 642 & 641.2 & \textbf{702} & \textbf{711} \\
Average Speed (m/s) & 3.46 & 5.07 & \textbf{5.6} & \textbf{6.01}  & 3.43 & 5.12 & \textbf{5.52} & \textbf{5.98} \\
Speed Violation (\%) & 0\% & 0\% & \textbf{0}\% & \textbf{0}\% & 0\% & 0\% & \textbf{0}\% & \textbf{0}\% \\
Crossing Duration (\%) & 77.19\% & 59.3\% & \textbf{70}\% & \textbf{48.89}\% & 61\% & 52.22\% & \textbf{60.2}\% & \textbf{48.88}\% \\
Average Stops & 11.51 & 7.01 & \textbf{6.11} & \textbf{6.11}  & 9.89 & 6.55 & \textbf{6.48} & \textbf{5.99} \\
Closet Pedestrian Distance (m) & 0.46 & 0.61 & \textbf{0.69} & \textbf{0.82}  & 0.41 & 0.6 & \textbf{0.67} & \textbf{0.75} \\
\midrule
\end{tabular}}
\caption{Performance comparison of all agents in known and unknown environments}
\label{results}
\end{table*}

%% file: conclusion.tex
\section{CONCLUSION}
This work formulates autonomous navigation in urban environments amonsgt pedestrians as a multi objective reinforcement learning problem. Two conflicting objectives are considered in this work, namely, \textit{safety} and \textit{speed}. Two variants of the proposed method are trained in a custom urban environment and are compared with similar single objective reinforcement learning agents in known and unknown environments.
All the agents are thoroughly compared  and evaluated and the results confirm the better performance of the proposed method.   

For future work, we plan to extend the method to incorporate more objectives such as considering traffic rules and passenger comfort. We also plan to extend the approach for a truly urban environment consisting of other vehicles along with pedestrians.